\documentclass[journal]{IEEEtai}

\usepackage[colorlinks,urlcolor=blue,linkcolor=blue,citecolor=blue]{hyperref}

\usepackage{color,array}

\usepackage{graphicx}

\usepackage{amsmath}
\usepackage[utf8]{inputenc} 
\usepackage[T1]{fontenc}    
\usepackage{hyperref}       
\usepackage{url}            
\usepackage{booktabs}       
\usepackage{amsfonts}       
\usepackage{nicefrac}       
\usepackage{microtype}      
\usepackage{xcolor}         
\usepackage{amssymb}
\usepackage{mathtools}
\usepackage{amsthm}

\usepackage{caption}
\usepackage{subcaption}

\begin{document}

\title{Hybrid Coordinate Descent for Efficient Neural Network Learning Using Line Search and Gradient Descent} 


\author{Yen-Che Hsiao and Abhishek Dutta
\thanks{Yen-Che Hsiao is with the Department of Electrical and Computer Engineering, University of Connecticut, Storrs CT 06269, USA. (e-mail: yen-che.hsiao@uconn.edu).}
\thanks{Abhishek Dutta is with the Department of Electrical and Computer Engineering, University of Connecticut, Storrs CT 06269, USA.}
}


\maketitle

\begin{abstract}
This paper presents a novel coordinate descent algorithm leveraging a combination of one-directional line search and gradient information for parameter updates for a squared error loss function. Each parameter undergoes updates determined by either the line search or gradient method, contingent upon whether the modulus of the gradient of the loss with respect to that parameter surpasses a predefined threshold. Notably, a larger threshold value enhances algorithmic efficiency. Despite the potentially slower nature of the line search method relative to gradient descent, its parallelizability facilitates computational time reduction. Experimental validation conducted on a 2-layer Rectified Linear Unit network with synthetic data elucidates the impact of hyperparameters on convergence rates and computational efficiency.
\end{abstract}

\begin{IEEEImpStatement}
This paper presents work whose goal is to advance the field of Machine Learning. There are many potential societal consequences of our work, none which we feel must be specifically highlighted here.
\end{IEEEImpStatement}


\section{Introduction}
\subsection{Problem Description}

We are given $n$ input-label training samples $S=\{\mathbf{X}_i,y_i\}_{i=1}^{n}$ where $\mathbf{X}_i \in \mathbb{R}^{p}$ represents the feature vector of the $i$-th data composed of $p$ features, $y_i \in \mathbb{R}$ denotes the corresponding labels.

We consider a two-layer neural network of the following form,
\begin{equation}
f(\mathbf{W},A,\vec{x}) = \frac{1}{\sqrt{m}}\sum_{r=1}^{m} a_r \sigma(\vec{w}_r^{T} \vec{x}),
\label{E1}
\end{equation}
where $m\in \mathbb{R}$ is the number of neurons, $\vec{x} \in \mathbb{R}^p$ is a feature vectors with $p$ features, $\vec{w}_r \in \mathbb{R}^p$ are the weight vectors in the first hidden layer connecting to the $r$-th neuron, $\mathbf{W}=\{\vec{w}_r\}_{r=1}^{m}$, $A=\{a_r\}_{r=1}^{m}$ is the set of weights in all other layers, $\sigma(x)=max\{x,0\}$ is the rectified linear unit (ReLU) activation function \cite{nair2010rectified}.

For this paper, we will use a combination of gradient descent and coordinate descent with Jacobi method \cite{bertsekas1997nonlinear} to minimize the mean square error (MSE) loss defined as 
\begin{equation}
L(\mathbf{W},A,\mathbf{X}) = \frac{1}{2}\sum_{i=1}^{n} (f(\mathbf{W},A,\mathbf{X}_i)-y_i)^2,
\label{E2}
\end{equation}
where $\mathbf{X}=\{\mathbf{X}_i\}_{i=1}^{n}$, and compare their performance in terms of convergence rate per epoch, computational complexity, and memory usage.

\subsection{Motivation}


Over-parameterized Deep Neural Networks (DNNs) have garnered considerable attention owing to their potent representation and broad generalization capabilities \cite{ergen2021convex}. In scenarios where the neural network is over-parameterized, the quantity of hidden neurons can greatly surpass either the input dimension or the number of training samples \cite{du2018power}. Despite classical learning paradigms issuing warnings against over-fitting, contemporary research indicates the prevalence of the "double descent" phenomenon, wherein substantial over-parameterization paradoxically enhances generalization \cite{liu2024aiming}.


A recent study has demonstrated that, in the case of two-layer neural networks, when learning is restricted locally within the neural tangent kernel space around particular initializations, neural networks behave like convex systems as the number of hidden neurons approaches infinity \cite{fang2022convex}. By leveraging the approximate convexity characteristic inherent in such neural networks, it may be possible to identify more effective optimizers (in contrast to the commonly employed gradient descent (GD) method) for training such neural networks.

\subsection{Related Work}



Several recent theoretical studies have revealed that in the case of exceedingly wide neural networks, wherein the number of hidden units scales polynomially with the size of the training data, the gradient descent algorithm, initiated from a random starting point, converges towards a global optimum \cite{oymak2020toward, zou2019improved}. 
Bu, Zhiqi et al. \cite{bu2021dynamical} demonstrate, from a dynamical systems perspective, that in over-parameterized neural networks, the Heavy Ball (HB) method converges to the global minimum of the mean squared error (MSE) at a linear rate similar to gradient descent, whereas the Nesterov accelerated gradient descent (NAG) may only achieve sublinear convergence.

\subsection{Major Challenges}


Recent research has demonstrated that an overparameterized neural network has the potential to converge to global optima under various conditions \cite{fang2021mathematical} or offer iteration complexity guarantees with different lower bounds on the number of neurons \cite{zou2019improved}. However, the majority of studies predominantly focus on utilizing gradient descent. With appropriate initialization, the training dynamics of over-parameterized DNNs during gradient descent can be characterized by a kernel function known as the neural tangent kernel (NTK) \cite{ye2024initialization} as outlined by Jacot et al. \cite{jacot2018neural}. Moreover, as the layer widths increase significantly, the Neural Tangent Kernel (NTK) converges to a deterministic limit that remains constant throughout training; therefore, in the scenario of infinite width, a key requirement for gradient descent to achieve zero loss is to ensure that the smallest eigenvalue of the NTK is kept away from zero \cite{bombari2022memorization}. A potential challenge lies in the fact that the methodology for analyzing neural network properties may vary depending on the optimization techniques employed. We aim to investigate whether other optimization methods can yield superior outcomes. 

\subsection{Proposed Method}


We experiment with the coordinate descent with Jacobi method outlined in Exercise 2.3.2 of the textbook by Bertsekas \cite{bertsekas1997nonlinear}. Let \(g: \mathbb{R}^n\rightarrow \mathbb{R}\) be a continuously differentiable convex function. For a given \(\vec{x}\in\mathbb{R}^n\) and all \(i=1,\dots,n\), define the vector \(\vec{x}^*\) by \({\vec{x}^*}_i\in \underset{\xi\in\mathbb{R}}{\mathrm{arg\,min}}\,g(\vec{x}_1,\dots,\vec{x}_{i-1},\xi,\vec{x}_{i+1},\dots,\vec{x}_n)\). The Jacobi method performs simultaneous steps along all coordinate directions, and is defined by the iteration 
\begin{equation}
\vec{x}:=\vec{x}+\alpha(\vec{x}^*-\vec{x}),
\label{E3}
\end{equation}
for \(\alpha\in\mathbb{R}\).


\section{Methods}

For our hybrid coordinate descent, the weights in the first layer are updated using (\ref{E3}) by
\begin{equation}
\vec{w}_r:=\vec{w}_r+\alpha({\vec{w}_r}^*-\vec{w}_r),
\label{E5}
\end{equation}
where \(\alpha=\frac{1}{n}\), \(r=1,\dots,m\), and \({\vec{w}_{r_i}}^*\in \underset{\xi\in\mathbb{R}}{\mathrm{arg\,min}}\,g(\vec{w}_{r_1},\dots,\vec{w}_{r_{i-1}},\xi,\vec{w}_{r_{i+1}},\dots,\vec{w}_{r_n})\).

We propose to use either the gradient of the loss with respect to the weight or coordinate descent to determine \({\vec{w}_r}^*\) in the Jacobi method in (\ref{E3}) for updating the parameters in the neural network defined in (\ref{E1}).

Each element in the vector \({\vec{w}_r}^*\) is updated using gradient by
\begin{equation}
{\vec{w}_{r_i}}^*:=\vec{w}_{r_1}-\frac{\partial L(\mathbf{W},A,\mathbf{X})}{\partial \vec{w}_{r_i}},
\label{E4}
\end{equation}
if the gradient \(\frac{\partial L(\mathbf{W},A,\mathbf{X})}{\partial \vec{w}_{r_i}}\) is larger than a threshold \(dw\in\mathbb{R}\).

If \(\frac{\partial L(\mathbf{W},A,\mathbf{X})}{\partial \vec{w}_{r_i}}<dw\), \({\vec{w}_{r_i}}^*\) is determined by a line search algorithm described as follows. First, we compute
\begin{equation}
L(\mathbf{W}_{r_i}^{+},A,\mathbf{X}) = \frac{1}{2}\sum_{i=1}^{n} (f(\mathbf{W}_{r_i}^{+},A,\mathbf{X}_i)-y_i)^2
\label{E_line_1}
\end{equation}
and
\begin{equation}
L(\mathbf{W}_{r_i}^{-},A,\mathbf{X}) = \frac{1}{2}\sum_{i=1}^{n} (f(\mathbf{W}_{r_i}^{-},A,\mathbf{X}_i)-y_i)^2,
\label{E_line_2}
\end{equation}
where \(\mathbf{W}_{r_i}^{+}=\{\vec{w}_1,\dots,\vec{w}_{i-1},\delta^+,\vec{w}_{i+1},\dots,\vec{w}_m\}\), \(\delta^+=\vec{w}_{i}+\epsilon\), \(\mathbf{W}_{r_i}^{-}=\{\vec{w}_1,\dots,\vec{w}_{i-1},\delta^-,\vec{w}_{i+1},\dots,\vec{w}_m\}\), \(\delta^-=\vec{w}_{i}-\epsilon\), and \(\epsilon=dw\). Next, if \(L(\mathbf{W}_{r_i}^{+},A,\mathbf{X})>L(\mathbf{W}_{r_i}^{-},A,\mathbf{X})\), we will keep decreasing \(\epsilon\) by \(\epsilon := \epsilon - dw\) and computing (\ref{E_line_2}) until the updated \(L(\mathbf{W}_{r_i}^{-},A,\mathbf{X})\) is less than the previous one. If \(L(\mathbf{W}_{r_i}^{+},A,\mathbf{X})<L(\mathbf{W}_{r_i}^{-},A,\mathbf{X})\), we will keep increasing \(\epsilon\) by \(\epsilon := \epsilon + dw\) and computing (\ref{E_line_1}) until the updated \(L(\mathbf{W}_{r_i}^{+},A,\mathbf{X})\) is less than the previous one. If (\ref{E_line_1}) is the same as (\ref{E_line_2}), we will iterate to the next weight variable.

\section{Experiment Results and Discussion}

In this section, we use synthetic data (similar to \cite{du2018gradient}) to compare the coordinate descent with Jacobi method and gradient descent and to demonstrate the effect of the hyperparameter, \(dw\), to the convergence rate and the computational complexity. We uniformly generate \(n=10\) data points from a \(d=1000\)-dimensional unit sphere and generate labels from a one-dimensional standard Gaussian distribution. Given a prediction network in (\ref{E1}), we initialize each weight vector in the first layer \(\vec{w}_r\) from a standard Gaussian distribution and initialize the weight in the last layer \(A\) from a uniform distribution over the interval \([-1,1]\).

\begin{figure*}[!t]
     \centering
     \includegraphics[width=\textwidth]{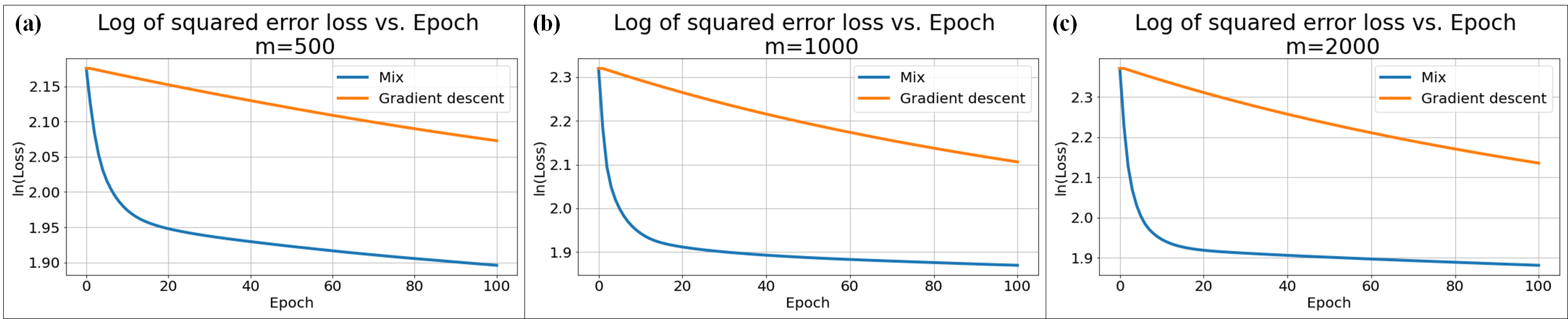}
        \caption{The natural logarithm of the loss with respect to epoch for the training of 2-layer ReLU network using gradient descent and our hybrid coordinate descent with various numbers of weight (\(m\)) in the first layer and with \(dw=0.5\).}
        \label{fig:1}
\end{figure*}

\begin{figure*}[!t]
     \centering
     \includegraphics[width=\textwidth]{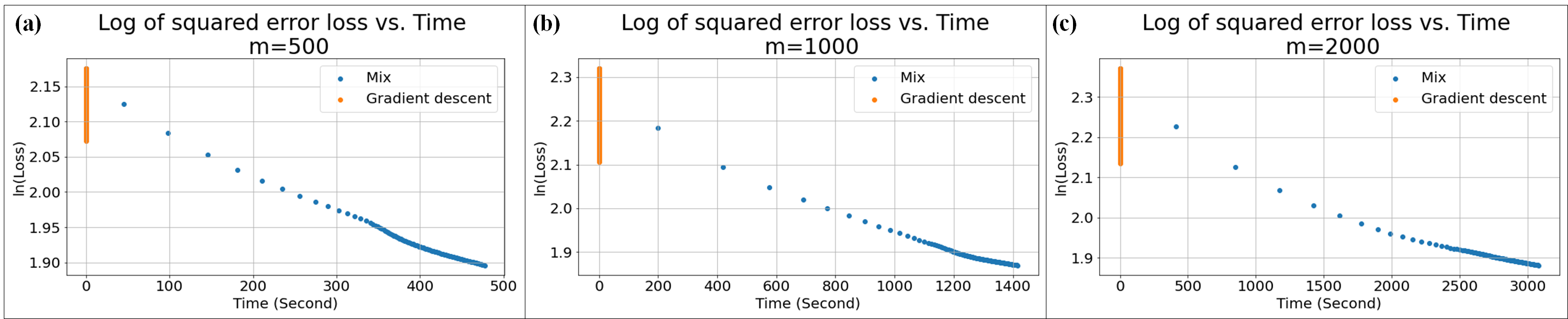}
        \caption{The natural logarithm with respect to the computational time (wall-clock time) for training of 2-layer ReLU network in 100 epochs using gradient descent and our hybrid coordinate descent with various numbers of weight (\(m\)) in the first layer and with \(dw=0.5\).}
        \label{fig:4}
\end{figure*}

\begin{figure*}[!t]
    \centering
    \includegraphics[width=\textwidth]{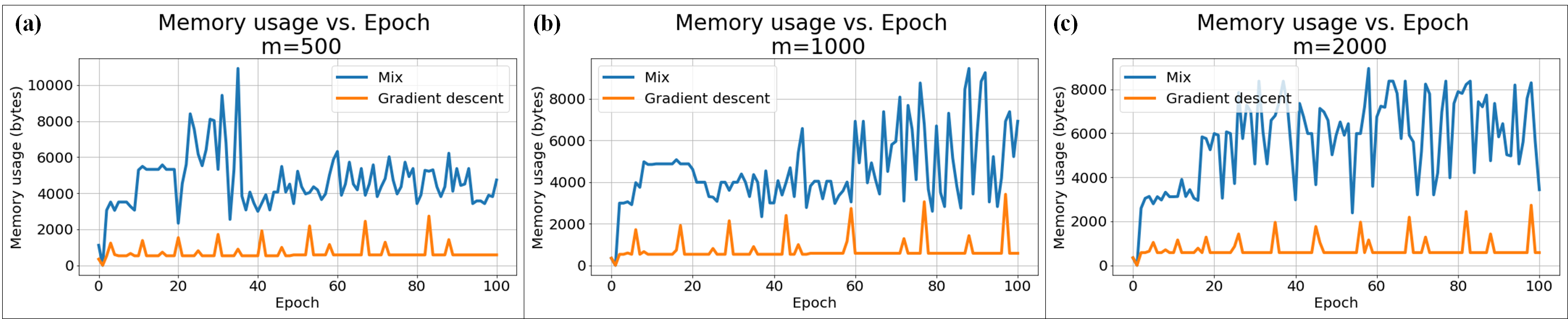}
    \caption{The natural logarithm of the loss with respect to epoch for the training of 2-layer ReLU network using gradient descent and our hybrid coordinate descent with various numbers of weight (\(m\)) in the first layer and with \(dw=0.5\).}
    \label{fig:6}
\end{figure*}

From Fig.~\ref{fig:1}, we can see that the convergence rate of our hybrid coordinate descent is faster than the gradient descent for the network with different number of hidden nodes. However, the comparison of the computational time between the two methods in Fig.~\ref{fig:4} shows that using gradient descent, the training of the 2-layer ReLU network for 100 epochs takes no more than the computational time of one epoch of our hybrid coordinate descent. In Fig.~\ref{fig:6}, we illustrate the memory usage of the random access memory for the two methods. It clearly demonstrates that the memory usage of the coordinate descent method is significantly higher than that of gradient descent.

\begin{figure*}[!t]
    \centering
    \includegraphics[width=\textwidth]{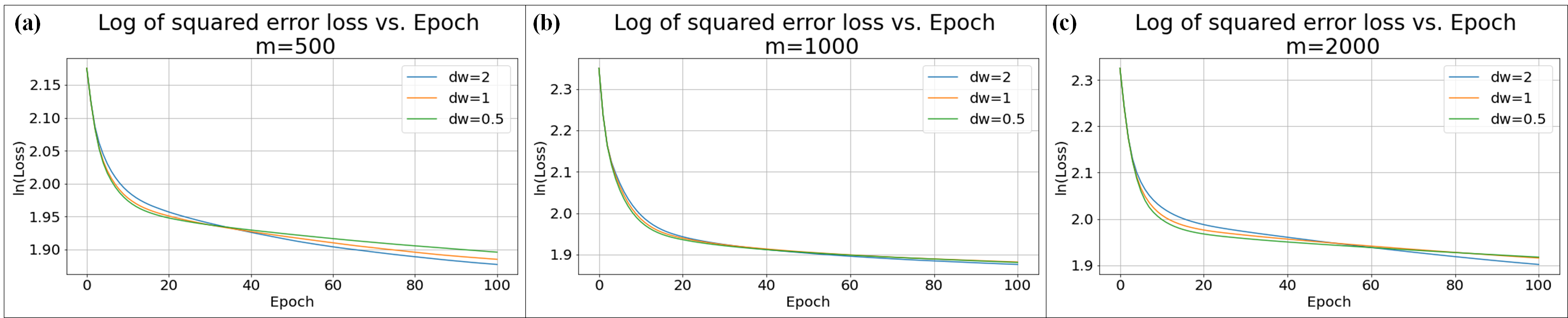}
    \caption{The natural logarithm of the loss with respect to epoch for the training of 2-layer ReLU network using our hybrid coordinate descent with various numbers of weight (\(m\)) in the first layer and with different value of \(dw\).}
    \label{fig:7}
\end{figure*}

\begin{figure*}[!t]
    \centering
    \includegraphics[width=\textwidth]{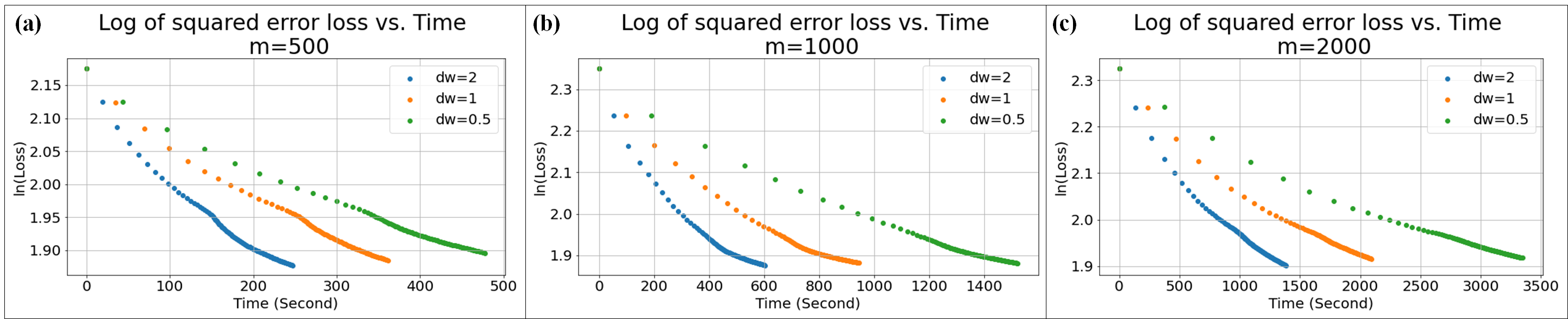}
    \caption{The natural logarithm of the loss with respect to time for the training of 2-layer ReLU network using our hybrid coordinate descent with various numbers of weight (\(m\)) in the first layer and with different value of \(dw\).}
    \label{fig:8}
\end{figure*}

Next, we show the effect of different threshold value, \(dw\), to the performance of training the ReLU network. In Fig.~\ref{fig:7}, it shows that with higher value of \(dw\), we can have a better convergence rate compare to the network that is trained with a smaller value of \(dw\). The reason behind this may be that as \(dw\) is higher, more of the parameters will be updated by the line search method and these parameters that are updated by the line search method may result in a lower loss value compared to the gradient method. As a result, our hybrid method with higher \(dw\) has a better convergence rate compared to that with a smaller \(dw\). In Fig.~\ref{fig:8}, it shows that with larger \(dw\), the computational time is smaller than the network trained with a smaller \(dw\). The reason may be that the step size of the line search algorithm with a higher \(dw\) is larger than the hybrid method with a smaller \(dw\). Therefore, the one with a higher \(dw\) can terminate the line search earlier than the one with a smaller \(dw\).

\section{Conclusion}

From the result, we can see that gradient descent is a better method than our hybrid coordinate descent in terms of computational time. However, if the computational time of our hybrid coordinate descent can be optimized to be close to that of in gradient descent, the convergence rate can be significantly improved since the loss decrease faster per epoch in the hybrid coordinate descent than gradient descent. Since gradient descent in PyTorch benefits from optimization in C++ code, while the implementation of the coordinate descent with Jacobi method is solely in Python, we think it is better to compare the two method without the C++ version of the gradient descent. In addition, the line minimization method we used may not be the best in terms of computational time. Thus, the future work of the study is to find out how to optimize the computational time in the hybrid coordinate descent or to see how to modify this method such that it can have a similar computational time to gradient descent. This could lead to faster convergence. Furthermore, since coordinate descent is designed for parallel computation, it is possible to implement parallelization and improve its speed. 





\bibliographystyle{ieeetr}
\bibliography{example_paper}

\end{document}